\documentclass{article}
\usepackage[numbers]{natbib}
\usepackage {arxiv}

\usepackage{cite}
\usepackage{floatrow}
\usepackage{amsmath,amssymb,amsfonts}
\usepackage{algorithm}
\usepackage{algorithmic}
\usepackage{textcomp}
\usepackage{xcolor}
\usepackage{float}
\usepackage{booktabs}
\usepackage{multirow}
\usepackage{enumitem}
\usepackage{setspace}
\usepackage{graphicx}
\usepackage{array}
\usepackage{amsmath}
\usepackage{amssymb}
\usepackage{epsfig}
\usepackage{epstopdf}
\usepackage{subfigure}
\usepackage{comment}
\usepackage{tabularx}
\usepackage{threeparttable}
\usepackage{adjustbox}
\usepackage{soul}
\usepackage{dsfont}
\usepackage[hyphens]{url}
\usepackage{hyperref} % href on \citepp & 

\title{\LARGE \bf
Ensemble of Unsupervised Deep Learning for Clustering Imbalanced Tabular Data}

\author{Pulock Das and Yina Hou\\
Department of Computer Science\\
Tennessee State University\\
Nashville, TN, USA\\
\and
{\bf 
Md. Kamrozzaman Bhuiyan}\\
 Enosis Solutions\\ 
 Dhaka, Bangladesh\\
\and
{\bf Manar D. Samad}\\
Department of Computer Science\\
North Carolina A\&T State University\\
Greensboro, NC, USA\\
\texttt{manar.samad@outlook.com} 
}

\begin{document}

\maketitle

\begin{abstract}

Data imbalance poses a major challenge in supervised classification, where the majority-class bias contributes to false negatives and overestimates classification accuracy. Unsupervised deep clustering can be immune to class imbalance because representation learning for clustering is performed without class labels. Deep clustering has been proposed for images, languages, and graphs, while its application to tabular data has only emerged recently. This paper is among the first to examine the performance of state-of-the-art deep clustering methods under varying levels of data imbalance. We introduce two novel cluster ensemble approaches: one aggregates deep clustering assignments across different embedding dimensions, and the other applies majority voting to the best-performing clustering algorithms. Experiments on 16 binary tabular datasets with varying and artificially induced levels of imbalance reveal distinct strengths of different deep clustering methods. On average, our ensemble methods outperform individual clustering methods in ACC, NMI, and ARI scores, offering greater resilience to data imbalance when identifying ground-truth classes without supervision. Therefore, in imbalance data settings, deep clustering can serve as a strong alternative to supervised classification.

\end{abstract}


\section{Introduction}

The deep learning revolution has largely centered on vision and language models, which have been repurposed for a vast range of supervised classification tasks. Supervised classification tasks are entirely dependent on labeled data in which class labels drive model training and performance. Model training supervised by class labels makes classifier models prone to overfitting and highly vulnerable to the well-known problem of class imbalance. Therefore, deep learning solutions for unsupervised data clustering can obviate the need for data labeling and may alleviate overfitting and data imbalance problems. Deep clustering approaches have been developed predominantly for image~\citep{caron2018deep}, language~\citep{xu2024text}, and graph data~\citep{liu2026survey}, largely overlooking one of the most prevalent data formats, namely tabular data~\citep{abrar2023effectiveness}. The clustering of tabular data still relies on traditional $K$-means clustering because a robust learning objective to train a deep neural network without involving class labels is not trivial. This paper investigates the performance of several newly proposed state-of-the-art deep clustering methods for tabular data with a three-fold objective. First, the paper evaluates and compares state-of-the-art deep clustering approaches, as these methods have not been systematically assessed against each other in earlier work. Second, we investigate the effectiveness of the proposed ensemble of deep clustering inferences in matching the actual class labels. Third, the state-of-the-art methods are evaluated at varying rates of data imbalance. The findings of this paper may provide useful directions for addressing major shortcomings of supervised classifiers in deep learning by effectively leveraging cutting-edge deep clustering methods.

% \subsection{Contributions}

\section{Related work}

Deep clustering methods primarily learn \emph{cluster-friendly} representations (embedding) for images \citep{xie2016unsupervised, guo2017improved} without using class labels. Clustering performance against the original class labels demonstrates the effectiveness of deep representation learning without class supervision. Various deep representation learning methods have been proposed for image datasets \citep{Boubekki2021, ghasedi2017deep, MoradiFard2020, Mrabah_neunet_2020} due to the underperformance of traditional clustering algorithms, such as K-means, in high-dimensional pixel spaces. In contrast, traditional clustering methods (e.g., K-means) are a \emph{de facto} choice for structured tabular data with raw features, whereas deep learning methods have not shown much success. However, K-means clustering performance is sensitive to the presence of uneven clusters~\citep{zhou2020effect}, which is equivalent to data imbalance in classification.  Recent deep clustering methods for tabular data \citep{rabbani2025deep, chen2024qgrl, svirsky2023interpretable, zhao2026tabclustpfn, vardakas2026deep} perform well on only a subset of datasets, and their robustness has not yet been systematically benchmarked to establish the state of the art.

Among recent methods, Gaussian Cluster Embedding in Autoencoder Latent Space (G-CEALS)~\citep{rabbani2025deep} learns the Gaussian distribution parameters of cluster embeddings in a deep learning framework. Despite competitive performance against traditional clustering, the performance of G-CEALS is sensitive to the embedding dimension. Quaternion Graph Representation Learning (QGRL) \citep{chen2024qgrl} uses heterogeneous data graphs to capture data relationships at different levels, such as between individual values, features, feature types, and samples. QGRL simultaneously optimizes a graph reconstruction and a spectral clustering objective to obtain a quaternion-based graph representation, which rotates feature vectors within a four-dimensional space. It outperforms several traditional and graph-based clustering methods on selective datasets. Interpretable Deep Clustering (IDC)~\citep{svirsky2023interpretable} is a tabular deep clustering method that learns cluster assignments leveraging feature-level explanations. A neural clustering head assigns samples to clusters by minimizing a coding-rate objective, which encourages well-separated clusters while keeping samples within each cluster tightly grouped. However, the performance of IDC is limited to high-dimensional biomedical tabular datasets. TabClustPFN~\citep{zhao2026tabclustpfn} extends a transformer-based tabular data classification framework, Tabular Prior-Data Fitted Network (TabPFN)~\citep{hollmann2025accurate}, to unsupervised clustering. TabClustPFN is pretrained in synthetic data clustering to subsequently cluster unseen tabular data in a single forward pass (zero-shot). Without requiring per-dataset retraining, TabClustPFN demonstrates state-of-the-art performance on synthetic and real-world tabular benchmarks. However, TabPFN methods can process a limited number of test samples due to memory limitations. Deep Clustering using the Soft Silhouette (DCSS) \citep{vardakas2026deep} optimizes a soft silhouette score using autoencoder-based deep clustering. Maximizing the silhouette score promotes compact and well-separated clusters. However, DCSS has so far been evaluated on only four tabular datasets. In addition, these existing methods rely on a fixed embedding dimension and do not systematically investigate the effect of this model hyperparameter on clustering performance.

This paper seeks to address three important gaps in the existing literature. First, individual deep clustering methods are generally evaluated on selective datasets. The literature does not report comparative performance of deep clustering methods for tabular datasets. Second, there is no single clustering algorithm that consistently achieves the best performance on every possible dataset. By aggregating the output of several clustering methods into an ensemble of cluster assignments, we may reduce the limitations of a single method. Third, it remains unknown how well deep clustering performs when faced with different degrees of class imbalance in tabular datasets. In deep representation learning, samples from the minority class can be absorbed into the majority class, leading to collapsed clusters within the latent space of an autoencoder~\citep{rabbani2025deep}. A clustering approach that shows robust performance under high class imbalances has an advantage over supervised classification techniques, which are strongly affected by this issue.

\section{Background}

\subsection{Preliminaries}
A tabular dataset $(X, y)$ consists of $N$ samples, each $D$-dimensional, where $X \in \mathbb{R}^{N \times D}$, and the corresponding class labels are denoted as $y \in \mathbb{Z}_{+}^{N}$. Given a total of $C$ distinct classes, a perfectly uniform class distribution is achieved when each class has $N_k = \frac{N}{C}$ samples, for $k = \{1, 2, \ldots, C\}$. Class imbalance occurs when there is substantial inequality in the class distribution, represented as \{$N_1 < N_2 < \ldots < N_C$\}, such that the size ratio between the majority class ($N_{mj}$) and the minority class ($N_{mn}$) can range from tenfold to one hundredfold. Some examples of highly imbalanced data scenarios include patient samples with a diagnosis of a particular disease and those without such a diagnosis in electronic health records~\citep{samad2026mining, hou2025causal}.  In digital communications, spam emails or messages linked to cyber-attacks are vastly outnumbered by legitimate messages~\citep{shanmugam2024addressing, shirvani2025advancing}. When one class outnumbers the others, a trained classifier can become biased, producing probability estimates that strongly favor this majority class. As a result, test instances from the minority class are often incorrectly classified.

\subsection{Existing solutions to data imbalance}

The problem of data imbalance is typically handled using one of two strategies. First, we adjust the overall class distribution by undersampling the majority class or oversampling the minority class until the classes are balanced. However, undersampling discards useful data points necessary for predictions~\citep{johnson2019survey, gao2026comprehensive}, whereas oversampling introduces artificial examples that may alter the original distribution of the data~\citep{chawla2002smote, johnson2019survey}. Second, the classifier is configured with class weights that assign a higher penalty to errors on minority-class samples~\citep{ling2008cost}. However, penalizing using class weights artificially forces the decision boundary, amplifies noise when many outliers can belong to the minority class, and is ineffective in multiclass imbalance~\citep{henning2023survey}. 

\section{Methodology}

\subsection {Proposed ensemble method}
We propose two strategies for an ensemble of deep clustering inferences. First, the G-CEALS method indicates that its performance depends on the specific selection of embedding dimensions used for clustering. We improve its clustering performance by aggregating clustering inferences made at different embedding dimensions as follows. We vary the dimensionality of the embedding from two to min (D, $R_{max}$), where D is the original feature dimension and $R_{max}$ is the maximum dimension of the embedding. For a given embedding dimension ($r_j$), we derive cluster assignments ($Y_{r_j}$) for all samples. It should be noted that, in the absence of ground-truth labels, the numerical identifiers associated with the clusters can be assigned arbitrarily. Randomly assigning cluster labels will introduce inconsistencies between the results of different embeddings, which in turn will degrade the quality of the ensemble results. Therefore, we select $r_j=2$ as the reference embedding for cluster assignments ($Y_{ref}$) and compare it with those from other embeddings. If the match in between is at least 50\% of the total labels, we retain the cluster assignments. Otherwise, the binary cluster assignments are flipped to ensure a concordance with the cluster assignment of the reference embedding. The adjustment of cluster assignments is shown in Equation~\ref{eq6_flip}.
%%%%%%%%%%%%%%%%%%
\begin{equation}
\hat{Y}(i,j) =
\begin{cases}
Y(i,j), & \text{if } \sum_{i} \mathbf{1}\{Y(i,j) == Y_{ref}(i)\} \geq N/2 \\[1mm]
1 - Y(i,j), & \text{otherwise}
\end{cases}
\label{eq6_flip}
\end{equation}
Here, Y(i, j) represents the cluster label of the $i$-th sample on the $j$-th embedding, given a total of N samples. After performing this adjustment, we obtain an ensemble cluster label for sample $i$ by averaging the cluster assignments across the embedding dimensions and then applying a threshold, as specified in the equations below.
%%%%%%%%%%%
\begin{eqnarray}
    Y_{avg} (i) &=& \frac{1}{d} \sum_{j=1}^{d} \hat{Y} (i, j) \\
    \label{eq:lavg}
    Y_{ens} (i) &=& \begin{cases}
  1, ~~ \mbox{if}~~Y_{avg} (i) \geq 0.5\\    
  0, ~~\mbox {otherwise}  
  \label{eq:lens}
\end{cases}
\end{eqnarray}
%%%%%%%%%%%%%
Second, we rank the clustering methods by overall performance and then obtain an ensemble of inferences from the top three methods using majority voting. The consistencies between the cluster assignments of the three methods are adjusted against a reference method (K-means clustering) using Equation~\ref{eq6_flip}.

\begin{table}[t]
\centering
\caption{Summary of two-class OpenML tabular datasets in this study. \emph{Total} represents the dimension including numerical (Num.) and categorical (Cat.) features.}
\label{tab:openml_datasets}
\centering
\scriptsize
\scalebox{1.0}{
\begin{tabular}{llrrrrll}
\toprule
ID & Dataset & Samples & Num. & Cat. & Total & Class ratio & Imbalance \\
\midrule
3     & kr-vs-kp      & 3196  & 0  & 36 & 73 & $1{:}1.1$  & Balanced \\
15    & breast-w      & 699   & 9  & 0  & 9  & $1{:}1.9$  & Balanced \\
31    & credit-g      & 1000  & 7  & 13 & 61 & $1{:}2.3$  & Mild \\
37    & diabetes      & 768   & 8  & 0  & 8  & $1{:}1.9$  & Balanced \\
44    & spambase      & 4601  & 57 & 0  & 57 & $1{:}1.5$  & Balanced \\
50    & tic-tac-toe   & 958   & 0  & 9  & 27 & $1{:}1.9$  & Balanced \\
151   & electricity   & 45312 & 7  & 1  & 14 & $1{:}1.4$  & Balanced \\
1049  & pc4           & 1458  & 37 & 0  & 37 & $1{:}7.2$  & High \\
1050  & pc3           & 1563  & 37 & 0  & 37 & $1{:}8.8$  & High \\
1063  & kc2           & 522   & 21 & 0  & 21 & $1{:}3.9$  & Moderate \\
1067  & kc1           & 2109  & 21 & 0  & 21 & $1{:}5.5$  & Moderate \\
1068  & pc1           & 1109  & 21 & 0  & 21 & $1{:}13.4$ & Very high \\
1464  & blood-trans   & 748   & 4  & 0  & 4  & $1{:}3.2$  & Moderate \\
1480  & Ilpd          & 583   & 9  & 1  & 11 & $1{:}2.5$  & Mild \\
1510  & wdbc          & 569   & 30 & 0  & 30 & $1{:}1.7$  & Balanced \\
40994 & climate       & 540   & 18 & 0  & 18 & $1{:}10.7$ & Very high \\
\bottomrule
\end{tabular}}

\end{table}

\subsection {Experiments on imbalanced data}

We conducted two experiments on binary tabular datasets to investigate the effects of class imbalance on clustering performance. First, the tabular datasets are grouped according to their class ratio, denoted as $CR$, into five class-imbalanced categories: balanced  $(CR<2)$, mild $(2\leq CR<3)$, moderate $(3\leq CR <7)$, high $(7 \leq CR <9)$, and very high $(CR \geq 9)$. Clustering performance is then evaluated by comparing the normalized results within each imbalance group. Second, we render five distinct class ratios (1:1, 1:5, 1:7, 1:10, and 1:15) to study the effect of varying degrees of class imbalance on clustering performance. In this rendering, the target class ratios are obtained by under-sampling the majority class to achieve a 1:1 ratio or by under-sampling the minority class for other ratios, relative to the size of the opposite class. Clustering performance is evaluated under both the original and controlled class ratios, always against the corresponding ground-truth class labels. All baseline clustering methods are obtained from the publicly available GitHub repositories of their original implementations.

\begin{table*}[t]
\centering
\caption{Clustering performance at varying imbalanced dataset groups. Scores are normalized across the methods and then averaged across the datasets for comparison. Bold and underline indicate the best and second-best scores, respectively. G-CEALS-EE is the ensemble-of-embeddings variant. KGT is the majority-voting ensemble of K-means, G-CEALS-EE, and TabClustPFN.}
\label{tab:imbalance_label_results}
\scriptsize
\setlength{\tabcolsep}{3pt}
\renewcommand{\arraystretch}{1.08}
\scalebox{0.9}{
\begin{tabular}{llcccccccc}
\toprule
\textbf{Metric} & \textbf{Label} & \textbf{K-means} & \textbf{QGRL} & \textbf{IDC} & \textbf{DCSS} & \textbf{TabClustPFN} & \textbf{KGT} & \textbf{G-CEALS} & \textbf{G-CEALS-EE} \\
\midrule
\multirow{6}{*}{ACC}
& Balanced  & 0.587 (0.42) & \textbf{0.857 (0.26)} & 0.525 (0.45) & 0.561 (0.35) & 0.464 (0.45) & 0.755 (0.35) & 0.255 (0.40) & \underline{0.797 (0.35)} \\
& Mild      & 0.759 (0.33) & 0.324 (0.09) & 0.379 (0.54) & 0.339 (0.27) & 0.442 (0.63) & \textbf{1.000 (0.00)} & 0.423 (0.00) & \underline{0.942 (0.08)} \\
& Moderate  & \underline{0.933 (0.07)} & 0.616 (0.04) & \textbf{1.000 (0.00)} & 0.313 (0.29) & 0.127 (0.22) & 0.914 (0.06) & 0.722 (0.08) & 0.852 (0.11) \\
& High      & \textbf{0.921 (0.11)} & 0.683 (0.45) & 0.065 (0.09) & 0.207 (0.29) & 0.378 (0.42) & 0.745 (0.07) & 0.671 (0.01) & \underline{0.773 (0.08)} \\
& Very High & 0.435 (0.61) & 0.654 (0.49) & \textbf{0.695 (0.43)} & 0.017 (0.02) & 0.335 (0.39) & 0.612 (0.00) & 0.502 (0.00) & \underline{0.658 (0.33)} \\
\cmidrule(lr){2-10}
& Avg. Rank & 3.2 (1.92) & 4.4 (2.70) & 4.4 (3.21) & 6.8 (1.10) & 6.4 (1.52) & \underline{2.8 (1.10)} & 5.6 (1.34) & \textbf{2.4 (0.89)} \\
\midrule
\multirow{6}{*}{NMI}
& Balanced  & 0.437 (0.40) & \underline{0.680 (0.39)} & 0.177 (0.30) & 0.279 (0.26) & \textbf{0.782 (0.28)} & 0.456 (0.37) & 0.098 (0.11) & 0.392 (0.49) \\
& Mild      & 0.353 (0.42) & 0.075 (0.11) & 0.212 (0.25) & 0.038 (0.05) & \textbf{0.989 (0.02)} & 0.273 (0.00) & 0.314 (0.00) & \underline{0.566 (0.61)} \\
& Moderate  & 0.349 (0.36) & \underline{0.583 (0.51)} & 0.394 (0.12) & 0.034 (0.04) & \textbf{1.000 (0.00)} & 0.408 (0.36) & 0.476 (0.67) & 0.335 (0.57) \\
& High      & 0.120 (0.09) & 0.143 (0.20) & 0.213 (0.04) & 0.254 (0.36) & \textbf{0.886 (0.16)} & 0.808 (0.11) & \underline{0.828 (0.24)} & 0.748 (0.24) \\
& Very High & 0.510 (0.69) & 0.259 (0.34) & 0.053 (0.07) & 0.150 (0.21) & 0.274 (0.35) & \underline{0.525 (0.00)} & 0.473 (0.00) & \textbf{0.794 (0.29)} \\
\cmidrule(lr){2-10}
& Avg. Rank & 4.8 (2.17) & 4.8 (2.59) & 6.4 (1.14) & 6.8 (1.30) & \textbf{1.8 (1.79)} & \underline{3.4 (1.14)} & 4.2 (2.28) & 3.8 (2.39) \\
\midrule
\multirow{6}{*}{ARI}
& Balanced  & 0.456 (0.40) & \textbf{0.739 (0.35)} & 0.187 (0.32) & 0.318 (0.29) & \underline{0.726 (0.31)} & 0.434 (0.32) & 0.003 (0.01) & 0.415 (0.48) \\
& Mild      & 0.081 (0.11) & 0.449 (0.48) & 0.404 (0.55) & 0.336 (0.48) & \underline{0.698 (0.41)} & 0.281 (0.00) & 0.344 (0.00) & \textbf{1.000 (0.00)} \\
& Moderate  & 0.644 (0.29) & 0.548 (0.39) & 0.618 (0.08) & 0.000 (0.00) & \underline{0.720 (0.26)} & \textbf{0.762 (0.37)} & 0.474 (0.63) & 0.464 (0.46) \\
& High      & 0.501 (0.13) & 0.061 (0.09) & 0.083 (0.12) & 0.156 (0.20) & 0.764 (0.33) & \textbf{0.939 (0.02)} & 0.857 (0.02) & \underline{0.935 (0.09)} \\
& Very High & 0.541 (0.65) & 0.315 (0.36) & 0.218 (0.21) & 0.000 (0.00) & 0.394 (0.43) & \underline{0.697 (0.00)} & 0.632 (0.00) & \textbf{0.825 (0.25)} \\
\cmidrule(lr){2-10}
& Avg. Rank & 4.6 (2.07) & 4.6 (2.70) & 5.8 (1.64) & 6.8 (1.10) & \textbf{3.0 (1.41)} & \underline{3.0 (2.55)} & 5.0 (2.12) & 3.2 (2.68) \\
\midrule 
& Overall Rank & 4.2 & 4.6 & 5.5 & 6.8 & \underline{3.7} & \textbf{3.1} & 4.9 & \textbf{3.1} \\
\bottomrule
\end{tabular}
}

\end{table*}

\subsection{Baseline methods}

We evaluate and compare eight state-of-the-art clustering methods, including our proposed ones. Baseline methods include K-means, QGRL, IDC, DCSS, TabClustPFN, and G-CEALS. Our ensemble solutions comprise two approaches, based on Equations 1, 2, and 3. The first method is an ensemble of clustering predictions based on the G-CEALS embeddings, which we call the G-CEALS ensemble of embeddings (G-CEALS-EE). The second method is an ensemble of clustering predictions that integrates K-means, G-CEALS-EE, and TabClustPFN, referred to as KGT.

\subsection {Cluster evaluation metrics}
Cluster performance is assessed by contrasting predicted cluster labels with true class labels in various class-imbalance scenarios using three metrics: clustering accuracy (ACC), normalized mutual information (NMI) and adjusted Rand index (ARI). Because clustering is an unsupervised task, class labels are employed solely for externally assessing cluster quality and are not used at any point during training or hyperparameter optimization.

\begin{itemize} 
    \item \textbf{Accuracy (ACC):} ACC determines the best correspondence between the predicted cluster labels and the ground-truth labels using the Hungarian algorithm~\citep{Kuhn1955}. It is computed as follows:  
%%%%%%%%%%%%%%
\begin{equation}
ACC  = \underset{n}{\operatorname{max}} \frac{\sum_{k=1}^{M} 1 \{Y_{true}^k = = n(Y_{pred}^k)\}}{M}
 \label{equation-ACC}
\end{equation}
%%%%%%%%%%%%%
Here, $M$ is the total number of samples, and $n(\cdot)$ denotes the optimal correspondence between predicted and true labels. ACC is calculated using the mapping that yields the maximum agreement between the true and predicted labels.

\item \textbf{Normalized Mutual Information (NMI)}: A high mutual information suggests stronger agreement between the ground-truth labels and the predicted clusters. A high NMI indicates that the cluster assignments involve less uncertainty. NMI values lie between 0 and 1, with 1 representing perfect agreement. We denote by $I(G,P)$ the mutual information between the ground-truth labels $G$ and the predicted clusters $P$. Mutual information is normalized by the mean of the corresponding entropies $H(G)$ and $H(P)$.

\begin{equation}
\text{NMI}(G, P) = \frac{I(G,P)}{mean ({H(G) , H(P)})}
\label{eq:nmi}
\end{equation}
Since NMI is based on mutual information and entropy, it is invariant to the numeric permutation of class or cluster labels~\citep{estevez2009normalized}. 
\item \textbf{Adjusted Rand Index (ARI)}: ARI measures the similarity between the ground-truth labels and predicted clusters based on pairwise sample agreement, while correcting for chance, as shown in Eq.~\ref{eq:eval_ari}. 
%%%%%%%%%
\begin{equation}
\label{eq:eval_ari}
     ARI = \frac{RI - \mathbb{E}(RI)}{max(RI) - \mathbb{E}(RI)}, 
     ~~~RI = \frac{TP + TN}{{}^{N}C_2}.
\end{equation}
%%%%
Here, the Rand index (RI) is calculated over all ${}^{N}C_2$ sample pairs for $N$ samples. A true positive (TP) pair occurs when both samples are matched at the predicted level $(p_i=p_j)$ and the ground-truth level $(g_i=g_j)$. A true negative (TN) pair arises when the two samples differ both at the predicted level $(p_i \neq p_j)$ and at the ground-truth level $(g_i \neq g_j)$. $\mathbb{E}(RI)$ represents the expected value of the RI. ARI ranges from -1 to 1: 1 indicates perfect agreement, values near 0 indicate random-level agreement, and negative values indicate discordant clustering. Higher ARI values reflect stronger concordance between predicted clusters and true labels. It is widely used to evaluate clustering, especially under random assignments and imbalanced class distributions~\citep{santos2009use}.

\end{itemize}

\begin{figure}[t]
    \centering
    \includegraphics[width=0.9\textwidth]{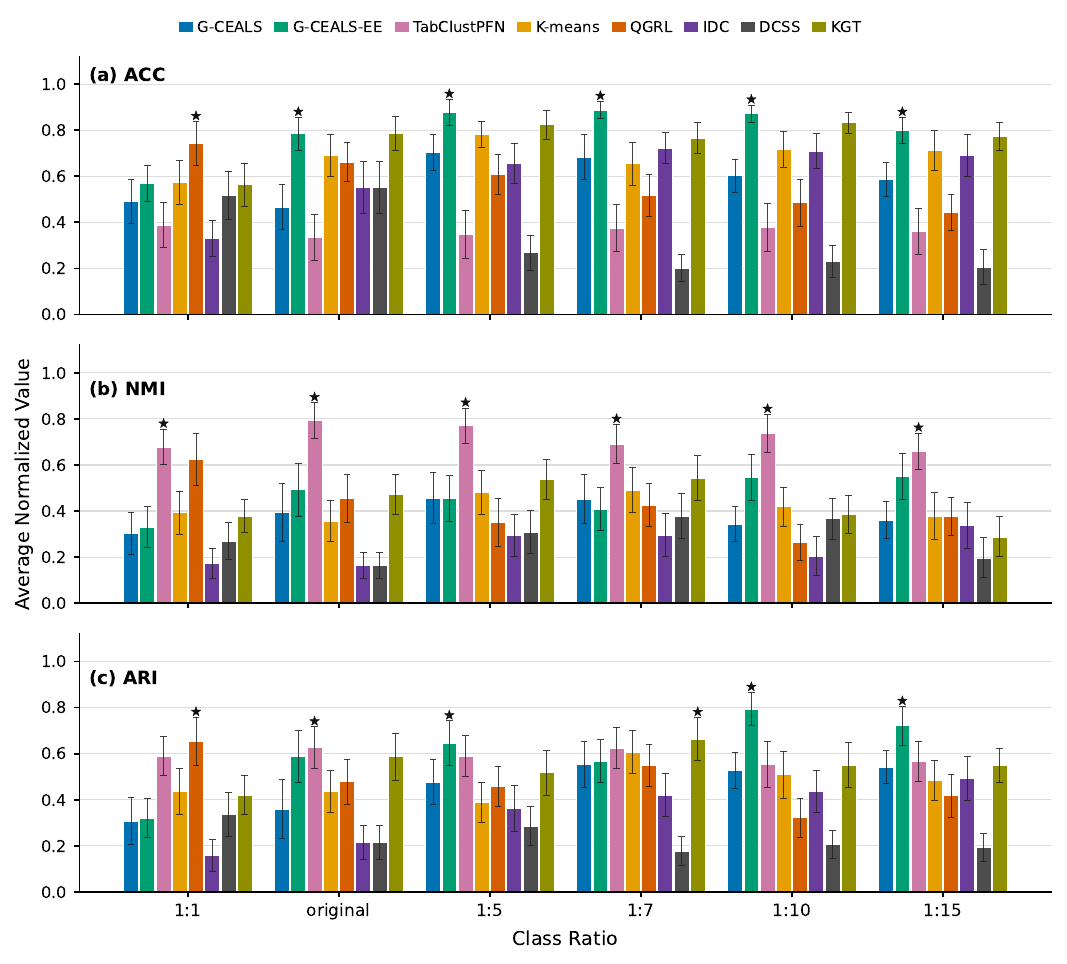}
    \vspace{-10pt}
    \caption{Normalized ACC, NMI, and ARI scores averaged across all datasets. Error bars denote the standard error. G-CEALS-EE is the ensemble-of-embeddings variant of G-CEALS. KGT is the ensemble of K-means, G-CEALS-EE, and TabClustPFN.}
    \label{fig:avg-value-combined}
\end{figure}

\section {Results}
Experiments have been conducted on four workstations, each with 64 GB RAM, using NVIDIA GPUs with 10–24 GB VRAM: RTX 3080 (10 GB), Quadro RTX 5000 (16 GB), RTX A4000 (16 GB), and RTX 4090 (24 GB). The embedding dimension of the G-CEALS-EE method is varied between 2 and min(33, D) in increments of 3, following a similar approach~\citep {samad2026mining}.

\subsection {Tabular datasets and imbalance settings}

Clustering methods are evaluated on 16 tabular datasets obtained from OpenML and summarized in Table~\ref{tab:openml_datasets}. The datasets cover a wide range of sample sizes and feature types, including numerical only, categorical only, and mixed feature types. In the binary classification setting, the class ratio varies between 1:1.1 (balanced) and 1:13.4 (highly imbalanced). For the other experiment, five additional controlled class-ratio configurations are created—$1{:}1$, $1{:}5$, $1{:}7$, $1{:}10$, and $1{:}15$—by randomly subsampling existing data only, without producing any synthetic samples. For the balanced setting, majority-class samples are downsampled to match the number of minority-class samples. For imbalanced settings, we keep all majority-class samples and downsample the minority class to match the target ratio. In general, to obtain a 1:N ratio against an original 1:M ratio, the minority class is downsampled when N$>$M, otherwise, we downsample the majority class.

%If the requested ratio requires more majority-class samples than available, all majority-class samples are kept and the minority class is downsampled instead. The final subset is then randomly shuffled using a fixed random seed of 42.

\subsection{Performance in imbalanced data groups}
The clustering performance metrics (ACC, ARI, NMI) are min–max normalized across all methods for each dataset, ensuring that the top-performing method attains a relative score of 1.0. The normalized scores are averaged within each dataset group according to the degree of data imbalance, as shown in Table~\ref{tab:imbalance_label_results}. QGRL performs strongly in the balanced group, achieving the best ACC and ARI scores and the second-best NMI score. However, its performance is very sensitive to clustering datasets with mild to very high imbalances. IDC shows a mixed response to class imbalance. It achieves the best ACC under moderate and very high data imbalance, but its NMI and ARI scores remain inconsistent and inferior in general. The differences in scores are most likely due to variations across datasets and the distinct, complementary insights each metric offers.

K-means follows a pattern similar to IDC, but attains a superior average rank across all three metrics. QGRL performance is on par with K-means, but K-means edges out on ACC scores. DCSS exhibits the poorest performance, possibly because it is optimized for the silhouette score, which may not correspond well with ACC, NMI, and ARI metrics. In other words, strong class separation does not guarantee effective class and cluster alignment. TabClustPFN is by far the best and most consistent clustering method based on NMI and ARI scores, but performs surprisingly poorly in terms of ACC. In general, the baseline G-CEALS performance is mediocre without optimization for embedding dimensions. In contrast, G-CEALS-EE consistently ranks among the best in ACC across all imbalance groups and achieves the highest NMI and ARI under very high imbalance. This may be due to trainable cluster weight parameter that regulates the cluster distribution. KGT attains the second-highest score across all three evaluation metrics. The average rank across ACC, NMI, and ARI indicates that the ensemble methods (KGT and G-CEALS-EE) achieve the best performance.

\subsection{Effects of varying imbalance ratios}

{Outcomes across different dataset groups may be affected by dataset-specific implicit biases. Therefore, in the second experiment, we explicitly render class imbalances at varying ratios. } In the balanced 1:1 setting, QGRL achieves the best clustering performance scores in general. Even with substantial undersampling, the clustering performance in the 1:1 setting still surpasses that achieved with the original class ratio. Compared to the balanced 1:1 setting, a 1:15 class ratio reduces ACC, NMI, and ARI by $40.3$\%, $39.3$\%, and $36.0$\%, respectively. Interestingly, IDC’s performance improves as imbalance increases. Higher clustering scores under imbalanced conditions than under the original class distribution suggest that class imbalance has a stronger effect than downsampling. 

% \pd{shouldn't we mention majority downsampling?}
% \mds {effect of downsample matters in general - we are losing samples, does not matter if they are from minority or majority}

K-means clustering exhibits a similar pattern; however, the influence of class imbalance is less pronounced than in IDC. The clustering effectiveness of DCSS deteriorates noticeably as the degree of class imbalance grows. TabClustPFN is the most resilient approach to class imbalance, achieving all the top NMI results and generally the second-best ARI results. In contrast to the other methods, its superior performance on the original class distribution compared with the undersampled variants indicates that undersampling can introduce some negative effects. Our ensemble embedding method, G-CEALS-EE, achieves the best overall ACC and ARI scores, along with several second-best NMI results. Its stable performance under higher imbalance ratios makes it particularly suitable for clustering imbalanced datasets. 
Figure~\ref{fig:avg-value-combined} illustrates the normalized ACC, NMI, and ARI scores of the individual clustering methods. The dominance of QGRL on balanced datasets, the consistently highest ACC scores of G-CEALS-EE, the consistently strongest NMI performance of TabClustPFN, and the mixed behavior observed in the ARI scores are in line with the findings reported in Table~\ref{tab:imbalance_label_results}.

\subsection{Summary of results}

Two experiments on data imbalance provide useful insight into the performance of state-of-the-art deep clustering methods, including the proposed ensemble solutions. The key findings of this paper can be summarized as follows. First, ensemble-based methods (G-CEALS-EE, KGT) deliver the most stable overall performance, mitigating the shortcomings of individual methods, as reflected in their best average ranks. Second, QGRL consistently stands out as the best clustering method for balanced datasets. Third, G-CEALS-EE yields markedly better clustering performance than the baseline G-CEALS, indicating that combining clustering results obtained from multiple embedding dimensions helps mitigate the uncertainties associated with relying on a single latent representation. G-CEALS-EE proves to be the most resilient method for handling highly imbalanced data. Fourth, optimizing silhouette scores for clustering, such as DCSS, does not guarantee better agreement with true labels. Last but not least, experiments on both naturally imbalanced data and synthetically induced imbalances reveal a consistent performance pattern, indicating that some methods handle data imbalance more robustly than others.

%\section{Discussion}
\section{Conclusion}

This study presents novel experiments on cutting-edge deep clustering methods to investigate their performance under varying degrees of data imbalance. Results reveal that ensemble-based clustering approaches most commonly provide the best performance under varying degrees of data imbalance. A limitation is that the proposed ensembles address binary clustering tasks to facilitate experiments on the data imbalance problem. Extending them to multi-class clustering would require a more general alignment strategy.

\bibliographystyle{elsarticle-num}
% \bibliography{New_Pruning}
\bibliography{ICHI26_cluster}

@article{Boubekki2021,
  title={Joint optimization of an autoencoder for clustering and embedding},
  author={Boubekki, Ahc{\`e}ne and Kampffmeyer, Michael and Brefeld, Ulf and Jenssen, Robert},
  journal={Machine learning},
  volume={110},
  number={7},
  pages={1901--1937},
  year={2021},
  publisher={Springer}
}

@inproceedings{ghasedi2017deep,
  title={Deep clustering via joint convolutional autoencoder embedding and relative entropy minimization},
  author={Ghasedi Dizaji, Kamran and Herandi, Amirhossein and Deng, Cheng and Cai, Weidong and Huang, Heng},
  booktitle={Proceedings of the IEEE international conference on computer vision},
  pages={5736--5745},
  year={2017}
}

@inproceedings{guo2017improved,
  title={Improved deep embedded clustering with local structure preservation.},
  author={Guo, Xifeng and Gao, Long and Liu, Xinwang and Yin, Jianping},
  booktitle={Ijcai},
  volume={17},
  pages={1753--1759},
  year={2017}
}

@article{kuhn1955,
  title={The Hungarian method for the assignment problem},
  author={Kuhn, Harold W},
  journal={Naval research logistics quarterly},
  volume={2},
  number={1-2},
  pages={83--97},
  year={1955},
  publisher={Wiley Online Library}
}

@article{MoradiFard2020,
  title={Deep k-means: Jointly clustering with k-means and learning representations},
  author={Fard, Maziar Moradi and Thonet, Thibaut and Gaussier, Eric},
  journal={Pattern Recognition Letters},
  volume={138},
  pages={185--192},
  year={2020},
  publisher={Elsevier}
}

@article{Mrabah_neunet_2020,
  title={Deep clustering with a dynamic autoencoder: From reconstruction towards centroids construction},
  author={Mrabah, Nairouz and Khan, Naimul Mefraz and Ksantini, Riadh and Lachiri, Zied},
  journal={Neural Networks},
  volume={130},
  pages={206--228},
  year={2020},
  publisher={Elsevier}
}

@article{estevez2009normalized,
  title={Normalized mutual information feature selection},
  author={Est{\'e}vez, Pablo A and Tesmer, Michel and Perez, Claudio A and Zurada, Jacek M},
  journal={IEEE Transactions on neural networks},
  volume={20},
  number={2},
  pages={189--201},
  year={2009},
  publisher={IEEE}
}

@article{rabbani2025deep,
  title={Deep clustering of tabular data by weighted Gaussian distribution learning},
  author={Rabbani, Shourav B and Medri, Ivan V and Samad, Manar D},
  journal={Neurocomputing},
  volume={623},
  pages={129359},
  year={2025},
  publisher={Elsevier}
}

@inproceedings{santos2009use,
  title={On the use of the adjusted rand index as a metric for evaluating supervised classification},
  author={Santos, Jorge M and Embrechts, Mark},
  booktitle={International conference on artificial neural networks},
  pages={175--184},
  year={2009},
  organization={Springer}
}

@inproceedings{xie2016unsupervised,
  title={Unsupervised deep embedding for clustering analysis},
  author={Xie, Junyuan and Girshick, Ross and Farhadi, Ali},
  booktitle={International conference on machine learning},
  pages={478--487},
  year={2016},
  organization={PMLR}
}

@inproceedings{chen2024qgrl,
  title={QGRL: quaternion graph representation learning for heterogeneous feature data clustering},
  author={Chen, Junyang and Ji, Yuzhu and Zou, Rong and Zhang, Yiqun and Cheung, Yiu-ming},
  booktitle={Proceedings of the 30th ACM SIGKDD conference on knowledge discovery and data mining},
  pages={297--306},
  year={2024}
}

@article{svirsky2023interpretable,
  title={Interpretable deep clustering for tabular data},
  author={Svirsky, Jonathan and Lindenbaum, Ofir},
  journal={arXiv preprint arXiv:2306.04785},
  year={2023}
}

@article{zhao2026tabclustpfn,
  title={TabClustPFN: A Prior-Fitted Network for Tabular Data Clustering},
  author={Zhao, Tianqi and Wang, Guanyang and Tan, Yan Shuo and Zhang, Qiong},
  journal={arXiv preprint arXiv:2601.21656},
  year={2026}
}

@article{vardakas2026deep,
  title={Deep clustering using the soft silhouette score: Towards compact and well-separated clusters},
  author={Vardakas, Georgios and Papakostas, Ioannis and Likas, Aristidis},
  journal={Machine Learning},
  volume={115},
  number={4},
  pages={81},
  year={2026},
  publisher={Springer}
}

@article{samad2026mining,
  title={Mining Electronic Health Records to Investigate Effectiveness of Ensemble Deep Clustering},
  author={Samad, Manar D and Hou, Yina and Ghosh, Shrabani},
  journal={arXiv preprint arXiv:2604.07085},
  year={2026}
}

@inproceedings{abrar2023effectiveness,
  title={Effectiveness of deep image embedding clustering methods on tabular data},
  author={Abrar, Sakib and Sekmen, Ali and Samad, Manar D},
  booktitle={2023 15th International Conference on Advanced Computational Intelligence (ICACI)},
  pages={1--7},
  year={2023},
  organization={IEEE}
}

@inproceedings{hou2025causal,
  title={Causal Explainability of Machine Learning in Heart Failure Prediction from Electronic Health Records},
  author={Hou, Yina and Rabbani, Shourav B and Hong, Liang and Diawara, Norou and Samad, Manar D},
  booktitle={2025 IEEE International Conference on Information Reuse and Integration and Data Science (IRI)},
  pages={128--134},
  year={2025},
  organization={IEEE}
}

@article{shanmugam2024addressing,
  title={Addressing class imbalance in intrusion detection: A comprehensive evaluation of machine learning approaches},
  author={Shanmugam, Vaishnavi and Razavi-Far, Roozbeh and Hallaji, Ehsan},
  journal={Electronics},
  volume={14},
  number={1},
  pages={69},
  year={2024},
  publisher={MDPI}
}

@article{shirvani2025advancing,
  title={Advancing Email Spam Detection: Leveraging Zero-Shot Learning and Large Language Models},
  author={Shirvani, Ghazaleh and Ghasemshirazi, Saeid},
  journal={arXiv preprint arXiv:2505.02362},
  year={2025}
}

@article{johnson2019survey,
  title={Survey on deep learning with class imbalance},
  author={Johnson, Justin M and Khoshgoftaar, Taghi M},
  journal={Journal of big data},
  volume={6},
  number={1},
  pages={27},
  year={2019},
  publisher={Springer}
}

@article{gao2026comprehensive,
  title={A comprehensive survey on imbalanced data learning},
  author={Gao, Xinyi and Xie, Dongting and Zhang, Yihang and Wang, Zhengren and Chen, Chong and He, Conghui and Yin, Hongzhi and Zhang, Wentao},
  journal={Frontiers of Computer Science},
  volume={20},
  number={11},
  pages={2011622},
  year={2026},
  publisher={Springer}
}

@article{chawla2002smote,
  title={SMOTE: synthetic minority over-sampling technique},
  author={Chawla, Nitesh V and Bowyer, Kevin W and Hall, Lawrence O and Kegelmeyer, W Philip},
  journal={Journal of artificial intelligence research},
  volume={16},
  pages={321--357},
  year={2002}
}

@article{ling2008cost,
  title={Cost-sensitive learning and the class imbalance problem},
  author={Ling, Charles X and Sheng, Victor S},
  journal={Encyclopedia of machine learning},
  volume={2011},
  number={2008},
  pages={231--235},
  year={2008},
  publisher={Springer Berlin, Germany}
}

@inproceedings{henning2023survey,
  title={A survey of methods for addressing class imbalance in deep-learning based natural language processing},
  author={Henning, Sophie and Beluch, William and Fraser, Alexander and Friedrich, Annemarie},
  booktitle={Proceedings of the 17th Conference of the European Chapter of the Association for Computational Linguistics},
  pages={523--540},
  year={2023}
}

@article{hollmann2025accurate,
  title={Accurate predictions on small data with a tabular foundation model},
  author={Hollmann, Noah and M{\"u}ller, Samuel and Purucker, Lennart and Krishnakumar, Arjun and K{\"o}rfer, Max and Hoo, Shi Bin and Schirrmeister, Robin Tibor and Hutter, Frank},
  journal={Nature},
  volume={637},
  number={8045},
  pages={319--326},
  year={2025},
  publisher={Nature Publishing Group UK London}
}

@inproceedings{caron2018deep,
  title={Deep clustering for unsupervised learning of visual features},
  author={Caron, Mathilde and Bojanowski, Piotr and Joulin, Armand and Douze, Matthijs},
  booktitle={Proceedings of the European conference on computer vision (ECCV)},
  pages={132--149},
  year={2018}
}

@article{xu2024text,
  title={Text clustering based on pre-trained models and autoencoders},
  author={Xu, Qiang and Gu, Hao and Ji, ShengWei},
  journal={Frontiers in Computational Neuroscience},
  volume={17},
  pages={1334436},
  year={2024},
  publisher={Frontiers Media SA}
}

@article{liu2026survey,
  title={A survey of deep graph clustering: Taxonomy, challenge, application, and open resource},
  author={Liu, Yue and Xia, Jun and Wu, Benyu and Zhou, Sihang and Yang, Xihong and Liang, Ke and Fan, Chenchen and Zhuang, Yan and Yu, Guoxian and Li, Stan Z and others},
  journal={IEEE Transactions on Knowledge and Data Engineering},
  year={2026},
  publisher={IEEE}
}

@article{zhou2020effect,
  title={Effect of cluster size distribution on clustering: a comparative study of k-means and fuzzy c-means clustering},
  author={Zhou, Kaile and Yang, Shanlin},
  journal={Pattern Analysis and Applications},
  volume={23},
  number={1},
  pages={455--466},
  year={2020},
  publisher={Springer}
}

\end{document}